\pdfoutput=1

\documentclass[11pt]{article}

\usepackage{emnlp2021}

\usepackage{times}
\usepackage{latexsym}

\usepackage[T1]{fontenc}

\usepackage[utf8]{inputenc}

\usepackage{microtype}

\usepackage{bm}

\usepackage{tabularx} 

\usepackage{graphicx}
\usepackage{subfig}

\usepackage[ruled,vlined]{algorithm2e}

\usepackage{multirow}

\usepackage{soul}
\usepackage{tikz}
\usetikzlibrary{calc}

\makeatletter
\newif\if@anonymize

\@anonymizefalse  

\if@anonymize
  \newcommand{\highlight@DoHighlight}{
    \fill [outer sep = -15pt, inner sep = 0pt, color=black]
          ($(begin highlight)+(0,8pt)$) rectangle ($(end highlight)+(0,-3pt)$) ;
  }

  \newcommand{\highlight@BeginHighlight}{
    \coordinate (begin highlight) at (0,0) ;
  }

  \newcommand{\highlight@EndHighlight}{
    \coordinate (end highlight) at (0,0) ;
  }

  \newdimen\highlight@previous
  \newdimen\highlight@current
  \newlength{\item@width}

  \DeclareRobustCommand*\anonymize{%
    \SOUL@setup
    \def\SOUL@preamble{%
      \begin{tikzpicture}[overlay, remember picture]
        \highlight@BeginHighlight
        \highlight@EndHighlight
      \end{tikzpicture}%
    }%
    \def\SOUL@postamble{%
      \begin{tikzpicture}[overlay, remember picture]
        \highlight@EndHighlight
        \highlight@DoHighlight
      \end{tikzpicture}%
    }%
    \def\SOUL@everyhyphen{%
      \discretionary{%
        \SOUL@setkern\SOUL@hyphkern
        \SOUL@sethyphenchar
        \tikz[overlay, remember picture] \highlight@EndHighlight ;%
      }{%
      }{%
        \SOUL@setkern\SOUL@charkern
      }%
    }%
    \def\SOUL@everyexhyphen##1{%
      \SOUL@setkern\SOUL@hyphkern
      \settowidth{\item@width}{##1}%
      \makebox[\item@width]{}%
      \discretionary{%
        \tikz[overlay, remember picture] \highlight@EndHighlight ;%
      }{%
      }{%
        \SOUL@setkern\SOUL@charkern
      }%
    }%
    \def\SOUL@everysyllable{%
      \begin{tikzpicture}[overlay, remember picture]
        \path let \p0 = (begin highlight), \p1 = (0,0) in \pgfextra
          \global\highlight@previous=\y0
          \global\highlight@current =\y1
        \endpgfextra (0,0) ;
        \ifdim\highlight@current < \highlight@previous
          \highlight@DoHighlight
          \highlight@BeginHighlight
        \fi
      \end{tikzpicture}%
      \settowidth{\item@width}{\the\SOUL@syllable}%
      \makebox[\item@width]{}%
      \tikz[overlay, remember picture] \highlight@EndHighlight ;%
    }%
    \SOUL@
  }
\else
  \newcommand{\anonymize}[1]{#1}
\fi
\makeatother
\title{EmoBERTa: Speaker-Aware Emotion Recognition in Conversation with RoBERTa}

\author{Taewoon Kim \and
Piek Vossen \\
Vrije Universiteit Amsterdam\\
\texttt{\{t.kim,p.t.j.m.vossen\}@vu.nl} \\
 }   

\begin{document}
\maketitle
\begin{abstract}
We present EmoBERTa: Speaker-Aware \textbf{Em}otion Recognition in Conversation with Ro\textbf{BERTa}, a simple yet expressive scheme of solving the ERC (emotion recognition in conversation) task. By simply prepending speaker names to utterances and inserting separation tokens between the utterances in a dialogue, EmoBERTa can learn intra- and inter- speaker states and context to predict the emotion of a current speaker, in an end-to-end manner. Our experiments show that we reach a new state of the art on the two popular ERC datasets using a basic and straight-forward approach. We've open sourced our code and models at \url{https://github.com/tae898/erc}.

\end{abstract}

\section{Introduction} \label{Introduction}

The scope of emotion recognition is very wide, ranging from stills of face images, audio data to the actual utterances or text as in tweets.
In this paper, we focus on emotion recognition in conversation (ERC), which is a subfield of emotion recognition. More specifically, the task is to predict the emotion of a current speaker who's engaging in a conversation with one person or more. Recognizing emotion is important to areas such as affective computing and human-robot communication, in which it can be an important feedback mechanism.

As humans use multiple sensory inputs to have a conversation (e.g., vision, voice, etc.) the ERC task can also include multiple modalities (e.g., visual, audio, text, etc.). Here, we report on our first experiments on the text modality, leaving using multiple modalities within our framework to future work.

Since the introduction of the Transformer \citep{NIPS2017_3f5ee243}, transformer-based deep neural network models have become the dominating neural network model in sequence modeling. Especially, pretrained encoder-only models such as BERT \citep{devlin-etal-2019-bert} and RoBERTa \citep{liu2019roberta} have shown that they can be successfully fine-tuned for downstream tasks, such as sentence classification and question answering. ERC can be seen as a special case of sequence modeling, since emotions are expected to be triggered by a preceding event in any modality.

Our approach to ERC enriches such transformer models by including the speaker identity in the sequence information spanning multiple utterances. By adapting the RoBERTa sequence representation, we improve the SOTA on two popular benchmark datasets.


\section{Related Work}
\label{sec:related}

Most of the existing works on ERC combine different kinds of neural network architectures (e.g., CNNs, RNNs, Transformers, GNNs, etc.) \citep{Li2021MultiTaskLW}, \citep{Li2020BiERUBE}, \citep{ishiwatari-etal-2020-relation}, \citep{wang-etal-2020-contextualized}, \citep{HAZARIKA20211}, \citep{sheng-etal-2020-summarize}, \citep{ghosal-etal-2020-cosmic}, \citep{DBLP:journals/corr/abs-1908-11540}. The biggest downside of such approaches is that each part of the model is responsible for extracting their own features. These extracted features might not be ideal for the other parts of the model. Also, since these models are combinations of sub-models, it is hard to understand what each sub-model is contributing and how to improve the overall model. Some of the approaches try to take advantage of external knowledge bases \citep{ghosal-etal-2020-cosmic}, \citep{DBLP:journals/corr/abs-1909-10681}, which adds even more complexity to the model.

A substantial number of approaches are heavily based on RNNs (e.g., GRU) to model the sequence \citep{DBLP:journals/corr/abs-1911-09075}, \citep{lu-etal-2020-iterative}. The biggest problem with this is that it inherently decouples the word embedding extraction and the sequence modeling, whereas BERT-like models tackle them at once, often leading to a better performance. Also, they have to rely on external decontextualized word embedding extractors (e.g., GloVe \citep{pennington-etal-2014-glove} or word2vec \citep{10.5555/2999792.2999959}). Furthermore, training an RNN is very inefficient since ``backpropagation through time'' has to wait until the last input of a sequence has been processed. 

The approaches that are most closely related to us are HiTrans \citep{li-etal-2020-hitrans} and DialogXL \citep{DBLP:journals/corr/abs-2012-08695}. HiTrans packs multiple utterances with \texttt{[CLS]} tokens prepended into one input sequence. This sequence is first fed into a BERT and then to another transformer. DialogXL is based on XLNet \citep{DBLP:journals/corr/abs-1906-08237}. Our approach differs in that we just use RoBERTa and encode the speaker information with multiple utterances.

\section{Methodology}
\label{sec:method}

\subsection{Problem Definition}

Let's say a dialogue of $M$ utterances is given and $I$ interlocutors are engaging in a conversation. Then the dialogue can be expressed as a list of vectors: $dialogue = [\bm{x_1}, \bm{x_2}, ..., \bm{x_M}]$,  where an utterance $\bm{x_t}$ contains several words (tokens). Each utterance $\bm{x_t}$ is spoken by a unique $ speaker \in$ $I$. Since this is a supervised setup, every utterance $\bm{x_t}$ has one corresponding label $y_t$ that is annotated by a human. 

The simplest way to solve this problem is to come up with a function $f$ that takes $\bm{x_t}$ as an input and outputs the correct label $y_t$. However, this doesn't take context into account. It's easily conceivable that the function $f$ should also consider the past utterances $[\bm{x_1}, \bm{x_2}, ..., \bm{x_{t-1}}]$ or even the future utterances $[\bm{x_{t+1}}, \bm{x_{t+2}}, ..., \bm{x_M}]$ to model the context.

\subsection{EmoBERTa}

EmoBERTa starts from the pretrained \texttt{roberta-large} model \citep{liu2019roberta}. Since the task is basically a sequence classification task, we simply add a randomly initialized linear layer with the softmax nonlinearity to the first hidden state (this state corresponds to the \texttt{[CLS]} token) of the last layer of the pretrained model. 

We chose RoBERTa, among the many BERT-like models, because its structure is not only relatively simple, but also it can deal with more than two segments. The original authors of RoBERTa simply used two \texttt{</s>} tokens consecutively as \texttt{[SEP]} token, which separates the first and the second segments. Although the pretrained model has not been trained on more than two segments, we show that EmoBERTa can be generalized to three segments per input sequence. 

The first, second, and third segments contain the past utterances, the current utterance, and the future utterances, respectively, in a dialogue. Each utterance is prepended with the name of a speaker so that the model is aware which utterance is spoken by whom. The task is to predict the emotion of the current utterance.

RoBERTa uses \texttt{<s>} and \texttt{</s>} as \texttt{[CLS]} and \texttt{[EOS]} tokens, respectively. Building an input sequence for EmoBERTa is outlined in Algorithm \ref{alg:input-construction}. Example sequences can be found in Figure \ref{fig:qualitative-analysis}.

\begin{algorithm}[h]
\SetAlgoLined

Given the current utterance $\bm{x_t} \in \{\bm{x_1}, \bm{x_2}, ..., \bm{x_M}\}$;
$max\_tokens = 512 - 2$; \\
$sequence = [SEP] + tokenize(\bm{x_t}) + [SEP]$; \\
$i = 1$; \\

\While{$len(sequence) <= max\_tokens$}{

Prepend $speaker(\bm{x_{t-i}}) + ``: " + \bm{x_{t-i}} $ to $sequence$;\\
Append $speaker(\bm{x_{t+i}}) + ``: " + \bm{x_{t+i}} $ to $sequence$;\\
$i = i + 1;$
}

Remove the last appended / prepended utterances; \\

$sequence = [CLS] + sequence + [EOS];$
\caption{Building an input sequence\protect\footnotemark}
\label{alg:input-construction}
\end{algorithm}

\footnotetext{The pretrained RoBERTa model can have a maximum of 512 tokens in one input sequence. The while loop terminates if there are no more available past / future utterances to add in the dialogue. We empirically found that capitalizing the names of the interlocutors leads to slightly better results.}

\subsection{Training}

The loss is calculated as the sum of cross entropy loss and $L^{2}$ weight decay \citep{10.5555/2986916.2987033}. We use adaptive gradient descent \citep{DBLP:journals/corr/KingmaB14}, \citep{DBLP:conf/iclr/LoshchilovH19} with gradual linear warmup learning rate scheduling \citep{DBLP:journals/corr/GoyalDGNWKTJH17}. The peak learning rate was determined using Optuna \citep{optuna_2019}. Mixed floating point precision was used to reduce the training time and increase the batch size \citep{DBLP:journals/corr/abs-1710-03740}. See Appendix \ref{sec:training-details} for the details (e.g., hyperparameters, training time, hardware, etc.). We mostly used the huggingface transformer pytorch library for training \citep{wolf-etal-2020-transformers}. 

Then the training loss function is

\begin{equation}
\mathcal{L}(\bm{w}) = -\frac{1}{N}\sum^{N}_{i=1}\sum^{C-1}_{c=0}y_{c}^{(i)}\log(\hat{y}_{c}^{(i)}) + \frac{\lambda}{2}\|\bm{w}\|_2^2
\label{eq:csl}
\end{equation}

where $\bm{y}_{}^{(i)}$ , is a one-hot label vector, $\hat{\bm{y}}_{}^{(i)}$ is the softmax output vector given the input ${\bm{x_{}^{(i)}}}$, $N$ is the number of training data samples, $C$ is the number of classes, $\lambda$ is a $L^2$ regularization rate, and $\bm{w}$ are the weights of the model. In practice, the data samples are batched, and stochastic gradient descent is used.

Although the weights $\bm{w}$ were tuned to minimize the loss value, the final model is chosen where the weighted $f_1$ score on the validation split is the highest, since that's the metric that we report.

\section{Experiments}
\label{sec:experiments}

\subsection{Datasets and Evaluation Metrics}

\begin{table}[h]
\centering
\resizebox{\columnwidth}{!}{\begin{tabular}{l|c|c|c}
\hline
        \multirow{2}{*}{Dataset}& \multicolumn{3}{c}{\textbf{number of dialogues (utterances)}}\\ 
        & train                     & val                    & test                   \\ \hline
MELD    & 1,038 (9,989)             & 114 (1,109)            & 280 (2,610)            \\ 
IEMOCAP & 100 (4,778)               & 20 (980)             & 31 (1,622)             \\ \hline \hline
        \multirow{2}{*}{Dataset}&\multicolumn{3}{c}{\textbf{mean number of utterances per dialogue (std.)}}\\
        & train                     & val                    & test                   \\ \hline
MELD    & 9.6 (5.8)                 & 9.7 (5.4)              & 9.3 (5.7)              \\ 
IEMOCAP & 47.78 (16.47)             & 49.0 (17.44)           & 52.32 (17.36)            \\ \hline
\end{tabular}}
\caption{The upper half of the table shows the number of dialogues and utterances of the datasets.\protect\footnotemark. The bottom half shows the mean and the standard deviation of the number of utterances per dialogue. As shown, IEMOCAP has about 5 times more utterances per dialogue than MELD.}
\label{tab:datasets}
\end{table}

\footnotetext{The IEMOCAP dataset does not officially have train, val, and test splits. Therefore, we follow the splits used by \citep{zhong-etal-2019-knowledge}, as these splits are widely used.}

We test EmoBERTa on the two popular ERC datasets.                              
\textbf{MELD} \citep{poria-etal-2019-meld} is a multimodal (visual, audio, and text) and multi-party (more than two interlocutors in a dialogue) conversational dataset. It was collected from the TV series \textit{Friends}. The seven emotions are \texttt{neutral}, \texttt{joy}, \texttt{surprise}, \texttt{anger}, \texttt{sadness}, \texttt{disgust}, and \texttt{fear}. Weighted $f_{1}$ score is used to evaluate the performance, as the class distribution is highly imbalanced. 
                                             
\textbf{IEMOCAP} \cite{Busso2008IEMOCAPIE} is a multimodal (visual, audio, and text) and dyadic (only two interlocutors in a dialogue) conversational dataset. Ten actors participated in the data collection. Although the original dataset contains 11 different emotions, only six of them are used for evaluation. They are \texttt{neutral}, \texttt{frustration}, \texttt{sadness}, \texttt{anger}, \texttt{excited}, and \texttt{happiness}. As MELD, the class distribution is highly imbalanced and thus a weighted $f_{1}$ score will be used for evaluation. Unlike MELD, IEMOCAP does not officially have the names of the speakers. Therefore, we gave each actor a random name. See Appendix \ref{sec:iemocap-speaker-names} for the details.

Some statistics on train, val, and test splits of both datasets can be found at Table \ref{tab:datasets}.

\subsection{Baselines}

We compare our model with the models we mentioned in Section \ref{sec:related}.

\section{Results and Analysis}
\label{sec:results}

\subsection{Quantitative Analysis}

Table \ref{tab:quantitative-analysis} shows the performance of our models and the baselines. Our models outperform the other models on both MELD and IEMOCAP.

\begin{table}[h]
\centering
\resizebox{\columnwidth}{!}{\begin{tabular}{ll|c|c}
\hline
\multicolumn{2}{l|}{Model}                                                  & MELD           & IEMOCAP        \\ \hline
\multicolumn{2}{l|}{BERT+MTL \citep{Li2021MultiTaskLW}}                     & 61.90          & -              \\ 
\multicolumn{2}{l|}{BiERU-lc \citep{Li2020BiERUBE}}                         & 60.84          & 64.65          \\ 
\multicolumn{2}{l|}{DialogueGCN \citep{DBLP:journals/corr/abs-1908-11540}}  & 58.1           & 64.18          \\ 
\multicolumn{2}{l|}{RGAT \citep{ishiwatari-etal-2020-relation}}             & 60.91          & 65.22          \\ 
\multicolumn{2}{l|}{CESTa \citep{wang-etal-2020-contextualized}}            & 58.36          & 67.1          \\
\multicolumn{2}{l|}{VHRED \citep{HAZARIKA20211}}                            & -              & 58.6          \\
\multicolumn{2}{l|}{SumAggGIN \citep{sheng-etal-2020-summarize}}            & 58.45          & 66.61          \\
\multicolumn{2}{l|}{COSMIC \citep{ghosal-etal-2020-cosmic}}                 & 65.21          & 65.28          \\
\multicolumn{2}{l|}{KET \citep{zhong-etal-2019-knowledge}}                  & 58.18          & 59.56          \\
\multicolumn{2}{l|}{BiF-AGRU \citep{DBLP:journals/corr/abs-1911-09075}}     & 58.1           & 63.5          \\
\multicolumn{2}{l|}{Iterative \citep{lu-etal-2020-iterative}}               & 60.72          & 64.37          \\
\multicolumn{2}{l|}{HiTrans \citep{li-etal-2020-hitrans}}                   & 61.94          & 64.5           \\ 
\multicolumn{2}{l|}{DialogXL \citep{DBLP:journals/corr/abs-2012-08695}}     & 62.41          & 65.94          \\ \hline

\multirow{4}{*}{\textbf{EmoBERTa}} & No past and future utterances          & 63.46          & 56.09          \\  
                                   & Only past utterances                   & 64.55          & \textbf{68.57} \\  
                                   & Only future utterances                 & 64.23          & 66.56          \\ 
                                   & Both past and future utterances        & \textbf{65.61} & 67.42          \\ 
                                   & $\,\to\,$ \textit{without speaker names} & 65.07        & 64.02         \\ \hline
\end{tabular}}
\caption{All of the reported values are weighted $f_1$ (\%) score on the test splits. The best model and the best performance values are in bold. Since the values are stochastic in nature, we report the mean values of five random seeds.}
\label{tab:quantitative-analysis}
\end{table}

EmoBERTa shows very good results: max weighted $f_1$ scores (\%) of $65.61$ (MELD) and $68.57$ (IEMOCAP) respectively and above the best reported SOTA, especially considering that no modifications were made to the original RoBERTa model architecture. We also trained a model without the speaker names prepended, which drops the performance: weighted $f_1$ scores (\%) of $65.07$ (MELD) and $64.02$ (IEMOCAP) respectively, providing evidence that encoding the speaker information helps.

Especially on IEMOCAP dataset, although we had to come up with random names for the actors, EmoBERTa was able to learn what's important. As for IEMOCAP, the results were better using only past utterances, rather than using both past and future utterances. We believe that this is due to the fact that IEMOCAP has many more utterances than MELD, and thus we couldn't fit all the past and future utterances in one sequence, meaning that only using past utterances can fit more past utterances than aiming for both. Past utterances were apparently more useful than future utterances to predict the emotion of a current utterance.

There was a bigger performance gain by incorporating past and/or future utterances in IEMOCAP than MELD ($12.48$ vs. $2.15$). We believe that this is due to the fact that the nature of IEMOCAP is more contextual than that of MELD.

\subsection{Qualitative Analysis}
 
To get more insight in the value of encoding the speaker, we did a qualitative analysis on 10 correctly and 10 incorrectly classified random samples from each test split. 

Our manual inspection shows that the model tries to learn the dynamics of the interlocutors in the beginning layer, as the current speaker tokens attend to the interlocutor tokens (We observed this behavior from all the 20 MELD and 20 IEMOCAP random test samples). 
As for MELD, in all 100\% of the correctly classified samples, based on the top 10 attended tokens, the \texttt{<s>} token of the last layer attended to the speaker token of the target (current) utterance, whereas this ratio was only 60\% for the incorrectly classified samples. This verifies that the current speaker token increasingly contains important information, as the tokens move on to the higher layers.

\begin{figure}[h]
  \centering
  \subfloat[A correctly classified example. Both the prediction and the truth are \texttt{joy}. ]{\includegraphics[width=0.45\textwidth]{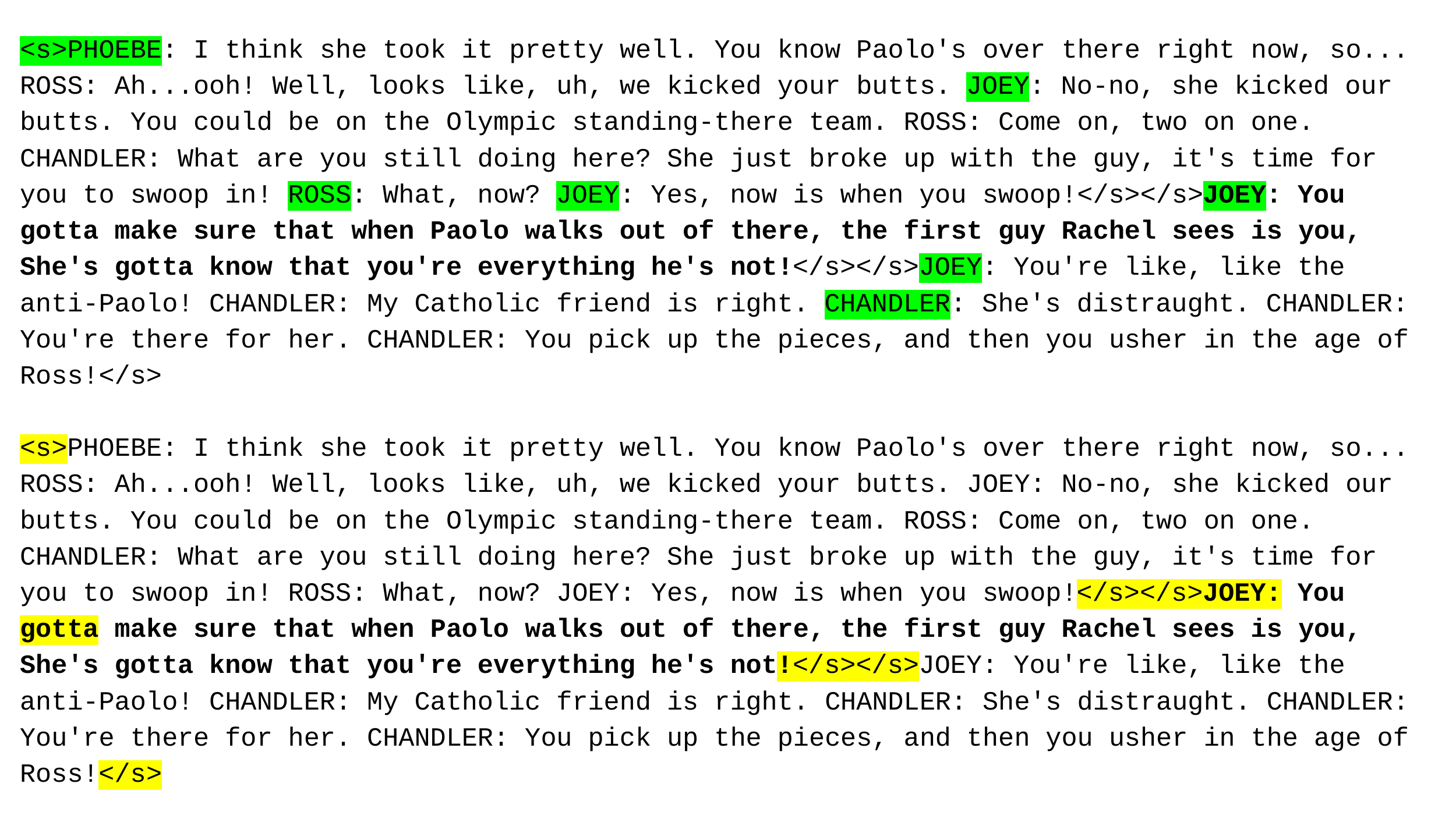}\label{fig:pred-joy_truth-joy}}
  \hfill
  \subfloat[An incorrectly classified example. The prediction is \texttt{joy} while the truth is \texttt{anger}.]{\includegraphics[width=0.45\textwidth]{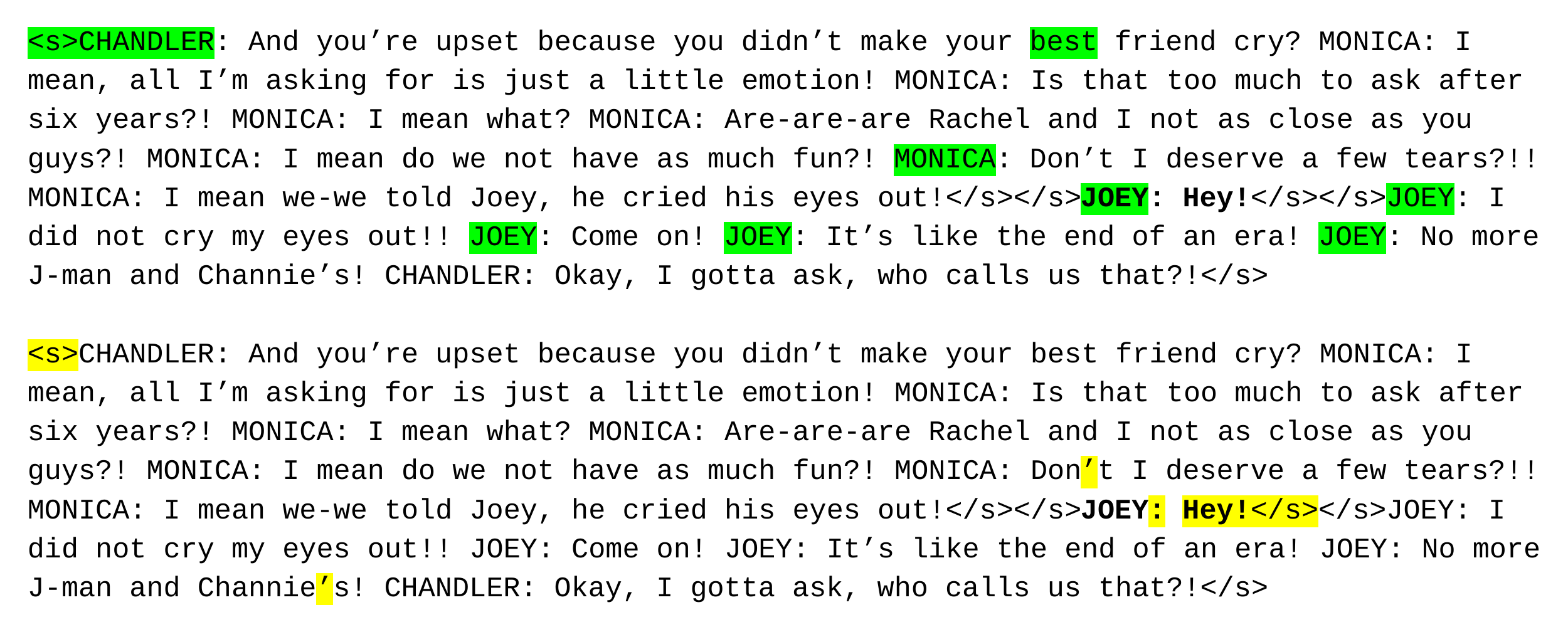}\label{fig:pred-joy_truth-anger}}
  \caption{Two examples from the 20 randomly selected test samples are shown. The current speaker utterance, of which the emotion that the model has to predict, is in bold. The green highlighted tokens are the top 10 most attended tokens to the current speaker (i.e., \texttt{JOEY}) in the beginning layer of the model. The yellow highlighted tokens are the top 10 most attended tokens to the \texttt{[CLS]} token (i.e., \texttt{<s>}) in the last layer.\protect\footnotemark. Best viewed when zoomed in.}
  \label{fig:qualitative-analysis}
\end{figure}

Figure \ref{fig:qualitative-analysis} shows a visualization of one correctly and one incorrectly classified random samples.

As the \texttt{<s>} token in the last layer focuses on the current speaker and his/her utterance, we believe that this information is what the model finds most useful to make the final prediction. 

Note that in the incorrectly classified example, the \texttt{<s>} token in the last layer does not focus on the current speaker but some random punctuation marks throughout the conversation, thus leading to an incorrect prediction.

See Appendix \ref{sec:qualitative-analysis-iemocap} for the random IEMOCAP test samples.

 \footnotetext{Since there are 16 attention heads used per layer in RoBERTa, the visualized weight coefficients are the mean values of them. Since RoBERTa uses a BPE tokenizer \citep{sennrich-etal-2016-neural}, the speaker names (e.g., \texttt{JOEY}) are often separated into more than one token (e.g., \texttt{JO} and \texttt{EY}). Therefore, we highlight the full names, even though only some parts of them are highlighted.}

\section{Conclusion}
\label{sec:conclusions}

In this paper, we showed that our new model, EmoBERTa, outperforms other models in the ERC task. Since EmoBERTa can directly attend to the input tokens and interlocutor names, we can easily observe the attention coefficients to see which part of the dialogue the model finds most important to make a final classification.

\section*{Acknowledgements}

\anonymize{This research was (partially) funded by the Hybrid Intelligence Center, a 10-year programme funded by the Dutch Ministry of Education, Culture and Science through the Netherlands Organisation for Scientific Research, https://hybrid-intelligence-centre.nl.}

\bibliography{anthology,custom}
\bibliographystyle{acl_natbib}

\clearpage
\appendix

\section{Appendix}
\label{sec:appendix}

\subsection{Training Details} \label{sec:training-details}

We used GCP (Google Cloud Platform) Compute Engine to carry out our experiments. We used an NVIDIA Tesla V100 machine (disclaimer: we are supported by neither Google nor NVIDIA). This GPU has 16 GB of memory and depending on the length of the input sequence, we were able to fit 4 to 16 samples in one batch, using mixed precision. The pretrained \texttt{roberta-large} model has about 355 million parameters. We found that without mixed precision, it's very difficult to train this model, since it's a pretty big model.

We set the value of $L^2$ regularization rate as 0.01. Training was done for five epochs. No weights were frozen during the training. The learning rate scheduler was set to linearly increase in the first 20\% of training and then linearly decrease in the remaining 80\%. 

Since the optimal peak learning rate highly depends on the batch size and the other hyperparameters, we used Optuna \citep{optuna_2019} to find its best value. 10\% of the training data and the same amount of validation data were used to search for the best value. Optuna ran five trials and looked for the best learning rate, between $1e-6$ and $1e-4$, that minimizes the cross entropy loss on the validation data split.

The hyperparameters not mentioned here are all set to the default values.

One full five-epoch training took about 45 minutes.

\subsection{IEMOCAP Speaker Names} \label{sec:iemocap-speaker-names}

Since the IEMOCAP dataset was created in the US, we used the top five male and female American names over the past 100 years (\url{https://www.ssa.gov/oact/babynames/decades/century.html}).

The female names used are \texttt{Mary}, \texttt{Patricia}, \texttt{Jennifer}, \texttt{Linda}, and \texttt{Elizabeth}. The male names used are \texttt{James}, \texttt{John}, \texttt{Robert}, \texttt{Michael}, and \texttt{William}.

\subsection{Qualitative Analysis on IEMOCAP} \label{sec:qualitative-analysis-iemocap}

Unlike MELD, only 20\% of the correctly classified samples showed the behavior of the last layer \texttt{<s>} token attending to the current target speaker. This ratio was 10\% for the incorrectly classified samples. We believe this is due to the fact that the speaker names in the test split of IEMOCAP (i.e., \texttt{WILLIAM} and \texttt{ELIZABETH}) were never seen during training.) 

\begin{figure}[h]
  \centering
  \subfloat[A correctly classified example. Both the prediction and the truth are \texttt{excited}.]{\includegraphics[width=0.45\textwidth]{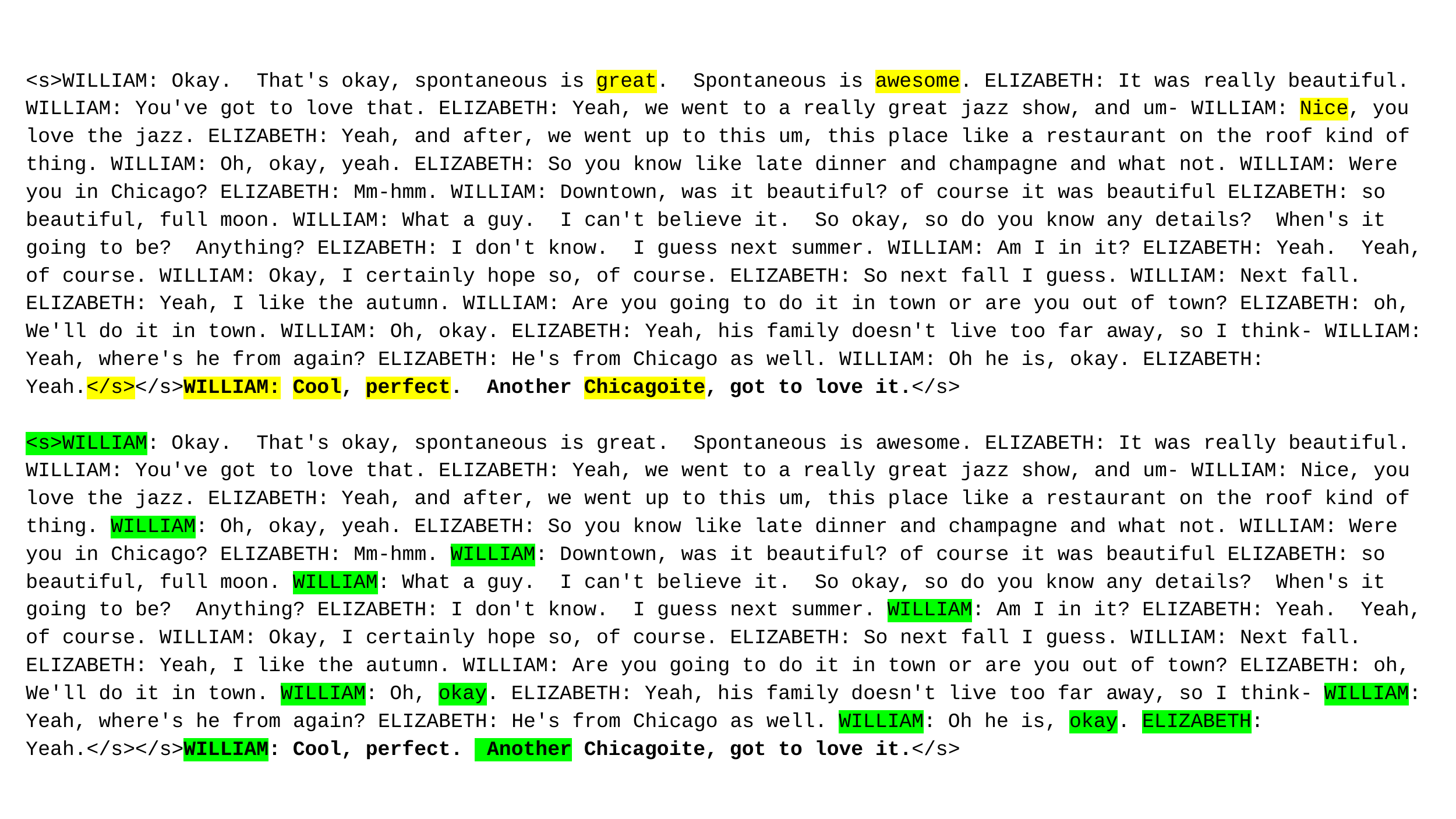}\label{fig:pred-excited_truth-excited}}
  \hfill
  \subfloat[An incorrectly classified example. The prediction is \texttt{neutral} while the truth is \texttt{frustration}.]{\includegraphics[width=0.45\textwidth]{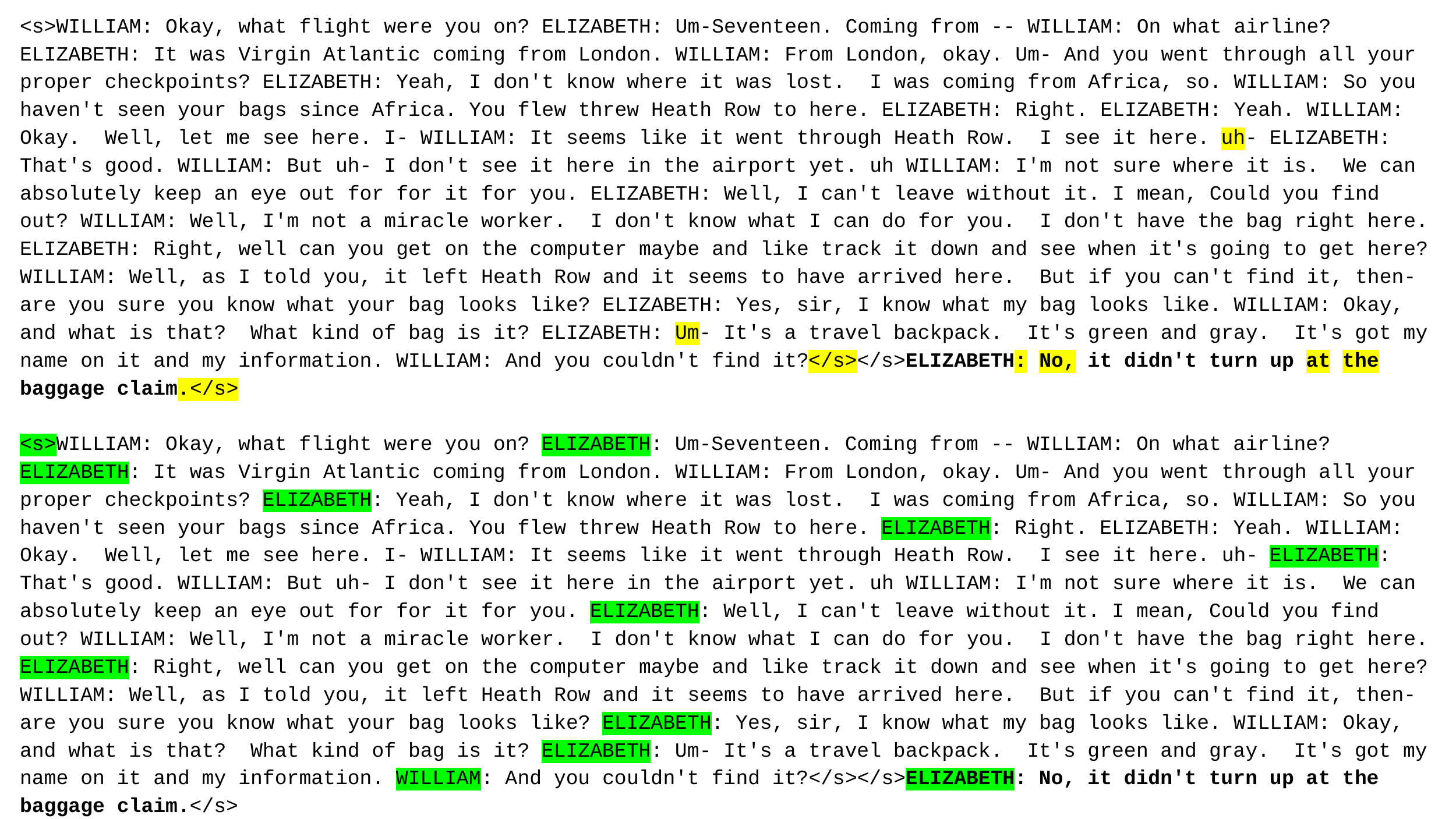}\label{fig:pred-neutral_truth-excited}}
  \caption{Two examples from the 20 randomly selected test samples are shown. The current speaker utterance, of which the emotion that the model has to predict, is in bold. The green highlighted tokens are the top 10 most attended tokens to the current speaker (i.e., \texttt{WILLIAM} and \texttt{ELIZABETH}, for Figure \ref{fig:pred-excited_truth-excited} and \ref{fig:pred-neutral_truth-excited}, respectively.) in the beginning layer of the model. The yellow highlighted tokens are the top 10 most attended tokens to the \texttt{[CLS]} token (i.e. \texttt{<s>}) in the last layer. Unlike Figure \ref{fig:qualitative-analysis}, there is only one \texttt{[SEP]} token (i.e., \texttt{</s></s>}), since this model only has two segments, past and current. Best viewed when zoomed in}
  \label{fig:qualitative-analysis-iemocap}
\end{figure}

Figure \ref{fig:qualitative-analysis-iemocap} gives you a visualization of the qualitative analysis on the IEMOCAP dataset.

We see a similar behavior as in MELD. Again, in the incorrectly classified example, the \texttt{<s>} in the last layer does not attend to the current speaker.

\end{document}